# Implementation of a "language driven" Backpropagation algorithm

I.V. Grossu[a,*], C.I. Ciuluvica (Neagu)[b,*]

[a] University of Bucharest, Faculty of Physics, Bucharest-Magurele, P.O. Box MG 11, 077125, Romania
[b] University of Bucharest, Faculty of Psychology and Education Sciences, 90 Panduri Street, District 5 Bucharest, P.O. 050663,  Romania

ABSTRACT

Inspired by the importance of both communication and feedback on errors in human learning, our main goal was to implement a similar mechanism in supervised learning of artificial neural networks. The starting point in our study was the observation that words should accompany the input vectors included in the training set, thus extending the ANN input space. This had as consequence the necessity to take into consideration a modified sigmoid activation function for neurons in the first hidden layer (in agreement with a specific MLP apartment structure), and also a modified version of the Backpropagation algorithm, which allows using of unspecified (null) desired output components. Following the belief that basic concepts should be tested on simple examples, the previous mentioned mechanism was applied on both the XOR problem and a didactic color case study. In this context, we noticed the interesting fact that the ANN was capable to categorize all desired input vectors in the absence of their corresponding words, even though the training set included only word accompanied inputs, in both positive and negative examples. Further analysis along applying this approach to more complex scenarios is currently in progress, as we consider the proposed "language-driven" algorithm might contribute to a better understanding of learning in humans, opening as well the possibility to create a specific category of artificial neural networks, with abstraction capabilities.

1. Introduction

Given the importance of communication in human learning [1-4], in this work we present an attempt of implementing a "communication language" in supervised learning of artificial neural networks. In this context, the significant role played by error feedbacks was noticed also. Thus, as Blair [1] suggests, nearly all categorization models take errors to be the essential ingredient in learning. According to Van Dyck et al. [5], the evaluation of past behaviors and acting upon the awareness that errors hold useful information are indeed considered as important practices to learn from errors. In literature, several authors: Edmonson [6], Van Dyck et al. [5], Rochlin [7], argue that communication is one of the most important conditions so that learning from errors should manifest.

In the field of artificial intelligence, communication in multi-agent systems [8] using neural networks is treated, for example, in the research of Steels & Belpaeme [9]. In the present work, we analyze a more restrictive, unidirectional ("trainer" - ANN) communication type, emphasizing the idea that words injected from the environment, in both positive and negative examples, should act as "learning triggers". Although we believe some of the discussed concepts could be used in the frame of other paradigms, for simplicity, this study was limited to the case of supervised learning.

*Corresponding authors. *E-mail addresses:*
*ioan.grossu@brahms.fizica.unibuc.ro (I.V.Grossu), cristina.ciuluvica@gmail.com (C.I.Ciuluvica)*

## 2. Implementation of a "language-supervised" training algorithm

Starting from the idea that both communication and feedback to errors should play an important role in supervised learning, our main purpose was to create an ANN with the following capabilities:

- Communicability. It is important to notice that we are treating only the particular case of unidirectional "trainer" - ANN communication, in which the language is used as a training instrument;
- Abstraction. Capability of recognizing words in the absence of any input vector;
- Pattern recognition. Capability of categorizing input vectors in the absence of corresponding language information;
- Sensitivity to errors. Capability of recognizing incorrect (word, input vector) associations.

Based on the observation that, in the training process, words should extend the ANN input space, acting as "learning triggers", the proposed model is constructed by analogy with the human brain, which neural activity is based on two fundamental inseparable events: excitation and inhibition. In the human brain there are two simple inhibitory circuits (feedback – for local network activity), and feed-forward (generated by the activity of others cortical regions), representing the fundamental building blocks of the cortical architecture [10]. According to the literature, the relationship between excitation and inhibition may be involved in various physiological functions, including increasing stability of cortical activity, preventing runaway excitation [11], regulating the neuronal output [12], and also the supervised learning of human behavior (the Pavlov's Conceptual Nervous System).

In this context, we considered a specific ANN modular architecture, comprising one word recognition component (which might be implemented using various specific systems [13-15]), and a Multilayer Perceptron, responsible for learning the specific function of interest. The MLP main module has an "apartment" structure (interconnected regions of adjacent hidden and output neurons), and receives as input both regular data and the output of the word recognition component. We considered a specific relation between words and neural apartments. Thus, presenting a word to the ANN will result in inhibiting (infinite dendrite weights) all apartments, excepting a corresponding one. In a dynamically approach, each time a new word is presented to the ANN input, a new empty apartment will be allocated accordingly.

For avoiding the complexity involved by lexical ambiguity [13], one can restrict the communication to a basic dictionary, containing only unambiguous words. In this case, a unique positive integer could be associated to each word (zero indicating the absence of language information). For simplicity, we will further use the term "word" for both words and for their corresponding unique identifiers. In this context, the previous described ANN could be replaced with an equivalent, simplified system, which includes only the apartment MLP module, and which input space is extended with the language component. The relation between words and apartments is resolved in the first hidden layer, by considering constant language dendrite weights (not modified during the training process), equaling their corresponding words. The word dendrite contributes also to the neuron activation:

$$y = f\left(t + w_{word} + \sum_{i=1}^{n} w_i x_i\right) \quad (1)$$

where *y* is the neuron output, *f* the activation function, *t* the neuron threshold, $w_{word}$ is the weight of the language dendrite, *n* the dimension of the input data, *w* the dendrites weights vector, and *x* is the input vector.

The apartment inhibition is achieved by implementing a modified sigmoid activation function for neurons in the first hidden layer:

$$f(x, x_{word}, w_{word}) = \begin{cases} 0, & x_{word} <> o \text{ and } x_{word} <> w_{word} \\ \frac{1}{1+e^{-x}}, & else \end{cases} \quad (2)$$

where $x_{word}$ is the word input, and w the weight of the word dendrite

Inspired by the importance of error feed-backs in human learning [1-7], we considered that the training set should include both positive and negative examples. For error recognition, one can envisage various approaches:

- Extending the output space by adding one neuron for each error category;
- Using bipolar neurons (*1* will indicate an affirmation, while *-1* a negation);
- Associate errors with the absence of any output signal (the "zero output" criteria).

As training technique we used a modified version of the Backpropagation [16-18] algorithm, capable of ignoring error signals coming from a specified set of output neurons. The neurons that must be ignored are identified, during the training process, by their corresponding null (unspecified) desired outputs. The main role of this mechanism is related to the creation of some privileged training neural channels (apartments) inside the ANN.

### 3. Application to the XOR problem

For testing the proposed approach, an object oriented three layers perceptron was implemented in Microsoft C# 2010. The input space was extended with the language dimension (which accepts only integer values). The activation functions for neurons in the first hidden layer were defined according to Eq.2, while all other neurons are using the classical sigmoid function. The previously discussed version of the Backpropagation algorithm ignores the error signals coming from output neurons with unspecified (null) expected values. Thus, the desired output was designed as a vector of nullable values [19]. A set of specific techniques [20] have been also considered for improving the efficiency of the training algorithm: the neural weights are initialized using random values in the *[-2/nd, 2/nd]* interval, *nd* representing the dendrites number for the corresponding neuron, the learning rate *c* is decreased in time following a *c/n* law (where n is a strictly positive integer, incremented after a specified number of iterations), and the elements of the training set are permuted after each iteration. A "shaking ANN weights" mechanism was also implemented for avoiding local minimum attractors.

The training set must contain both positive and negative examples. In this context, the previously discussed "zero output" error recognition criterion was empirically chosen. As a training success condition, a maximum target error, specified as parameter, was required to be satisfied by each input vector from the training set. The classification error is defined based on the Euclidian distance between the ANN response and the desired output:

$$D_\varepsilon = 0.5 \sum_{i=0}^{n} (o_i - d_i)^2 \quad (3)$$

where *d* is the desired output, nulls being ignored, and *o* the corresponding output.

Following the belief that basic concepts should be tested on simple examples, the XOR problem has been chosen as a first concrete application to be discussed. A two-apartment MLP was used following this purpose (Fig.1.). As the number of effective patterns (word – input vector combinations) is increased with respect to the classical XOR problem, a higher number of neurons should be allocated in the hidden layer [21]. The training set, together with one approximate solution (*c = 0.5*, and *target error = 0.001*) is presented, for exemplification, in Tab.1.

| $x_1$ | $x_2$ | $x_{word}$ | $d_1$ | $o_1$ | $d_2$ | $o_2$ | $D_\varepsilon$ |
|---|---|---|---|---|---|---|---|
| \multicolumn{8}{c}{positive examples} |
| 0 | 0 | 1 | 1 | 0.96 | null | 0.45 | 0.0009 |
| 1 | 1 | 1 | 1 | 0.97 | null | 0.47 | 0.0004 |
| 0 | 1 | 2 | null | 0.09 | 1 | 0.96 | 0.0005 |
| 1 | 0 | 2 | null | 0.14 | 1 | 0.96 | 0.0007 |
| \multicolumn{8}{c}{negative examples} |
| 0 | 0 | 2 | null | 0.13 | 0 | 0.02 | 0.0002 |
| 1 | 1 | 2 | null | 0.11 | 0 | 0.04 | 0.001 |
| 0 | 1 | 1 | 0 | 0.04 | null | 0.49 | 0.0007 |
| 1 | 0 | 1 | 0 | 0.04 | null | 0.39 | 0.0008 |
| \multicolumn{8}{c}{result (not included in the training set)} |
| 0 | 0 | 0 | - | 0.96 | - | 0.02 | - |
| 1 | 1 | 0 | - | 0.96 | - | 0.05 | - |
| 0 | 1 | 0 | - | 0.04 | - | 0.95 | - |
| 1 | 0 | 0 | - | 0.03 | - | 0.97 | - |

**Tab.1.** An approximate solution for the XOR problem.

One can notice that the ANN is able to learn all examples from the training set, but it is also capable of categorizing all input vectors in the absence of their corresponding words, even though this kind of information was not explicitly included in the training set. The success rate, measured for 1000 distinct training processes, is 98%. For the example presented in Tab.2, it is also interesting to remark that, for all null desired values, the corresponding outputs are close to zero. However, this effect is obtained with a low frequency (19%). For avoiding this uncertainty, the algorithm was tested also with the complete training set (which includes input without word examples, and in which all nulls are replaced with zero). The same success rate was obtained in this case.

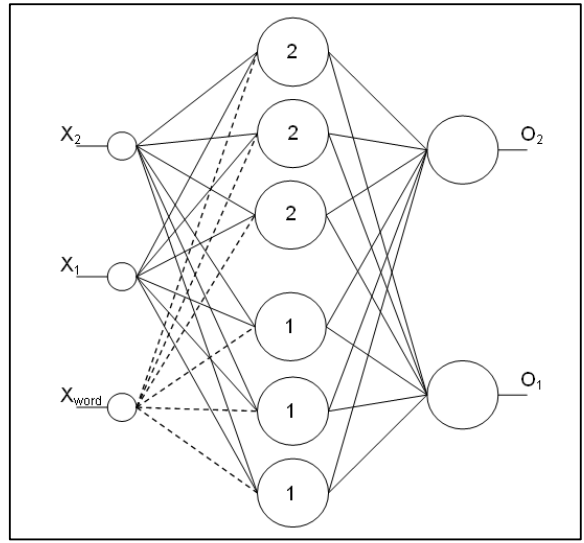

**Fig.1.** The ANN topology used for the XOR problem.

## 4. A didactic color case study

As a more complex example, we considered the didactic problem of categorizing the colors corresponding to the RGB cube vertices. An eight-apartment structure was used in this case. The number of hidden neurons per apartment (*12*) was empirically established. For each class, a number of 29 input vectors were symmetrically chosen around each vertex (Fig.2.). The training set is constructed by analogy with the data presented in Tab.1. It contains both positive and negative examples, summarizing *1856* elements. For avoiding high execution times, the target error was set to *0.1*, while the learning rate was empirically chosen (*c = 0.005*).

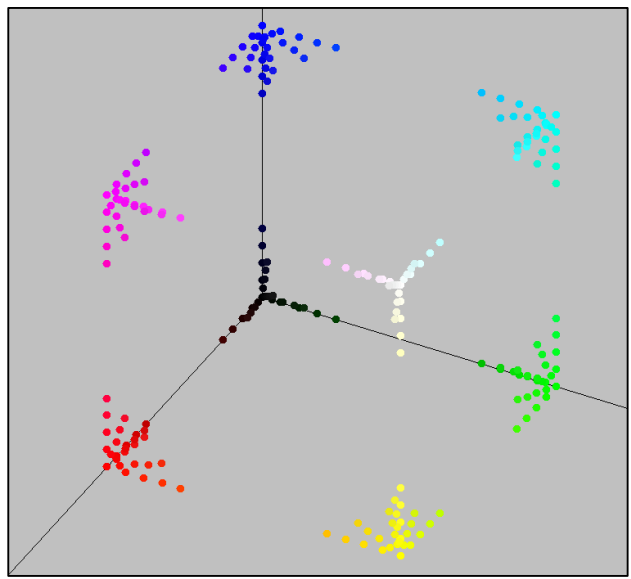

**Fig.2.** The RGB vectors used in the training set.

Similarly with the, previously discussed, XOR problem, the ANN is capable of categorizing all inputs without word vectors (Tab.2), even though this kind of information is missing from the training set.

| $x_{Red}$ | $x_{Green}$ | $x_{Blue}$ | $o_{Black}$ | $o_{Blue}$ | $o_{Magenta}$ | $o_{Red}$ | $o_{Yellow}$ | $o_{Green}$ | $o_{Cyan}$ | $o_{White}$ |
|---|---|---|---|---|---|---|---|---|---|---|
| 0 | 0 | 0 | 0.91 | 0.06 | 0 | 0.04 | 0.08 | 0.05 | 0 | 0.05 |
| 255 | 0 | 0 | 0.1 | 0 | 0.06 | 0.94 | 0.07 | 0.02 | 0.01 | 0.03 |
| 0 | 255 | 0 | 0.1 | 0.01 | 0 | 0.01 | 0.06 | 0.94 | 0.04 | 0.04 |
| 0 | 0 | 255 | 0.05 | 0.88 | 0.05 | 0.02 | 0 | 0.02 | 0.13 | 0.04 |
| 255 | 255 | 0 | 0.03 | 0 | 0.01 | 0.01 | 0.92 | 0.03 | 0 | 0.04 |
| 0 | 255 | 255 | 0.04 | 0.01 | 0 | 0.01 | 0 | 0.02 | 0.87 | 0.05 |
| 255 | 0 | 255 | 0.04 | 0.01 | 0.9 | 0.02 | 0.02 | 0.01 | 0.01 | 0.04 |
| 255 | 255 | 255 | 0.05 | 0.01 | 0 | 0.02 | 0.01 | 0.01 | 0.01 | 0.89 |

**Tab.2.** An approximate solution for the discussed RGB color categorization problem.

The algorithm success rate, measured for 300 distinct training processes, is 99.6%. We noticed also that, the output values for neurons without expectations (null desired value) have an accumulation to zero tendency (the absolute maximum value is 0.89, while, in average, 71% of all outputs are below 0.5, which is convenient for the target error we chose). The probability distribution of the maximum output for neurons with unspecified desired values is presented in Fig.3. Thus, the zero output error recognition condition could be extended to a more practical, "noisy output criterion" (high entropy output, for which a maximum value cannot be clearly distinguished). As opposed to the XOR problem, the algorithm did not prove its efficiency for the complete training set. One possible amelioration could come from using a robust Backpropagation algorithm [22], together with a more advanced local minimum avoiding technique (e.g. the simulated annealing [23]).

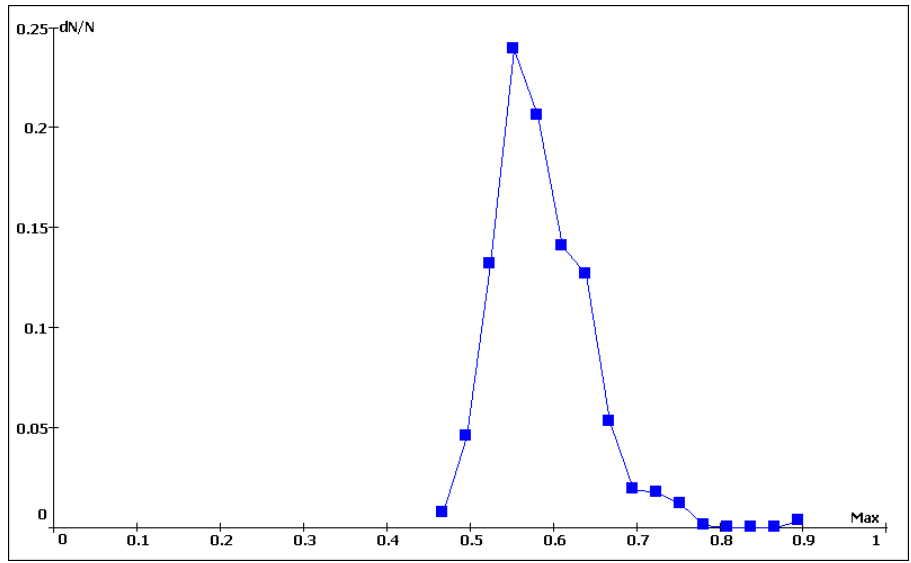

**Fig.3.** The probability distribution of the maximum output for neurons with unspecified desired values

## 5. Conclusions

Starting from the importance of both communication and feedback on errors in human learning, we tried to implement similar concepts in supervised learning of artificial neural networks. Following this purpose, the input space was extended with the language dimension, a modified sigmoid activation function was considered for neurons in the first hidden layer (in agreement with a specific MLP apartment structure), a modified Backpropagation algorithm was implemented, in order to allow using unspecified (null) desired output components, and both positive and negative examples were included in the training set. For a better understanding of basic concepts involved, the previous discussed mechanism have been tested on both the XOR problem and a simple RGB color case study. As an interesting result, one can notice that the ANN is capable of recognizing input vectors in the absence of their corresponding words, even when this kind of information is not explicitly included in the training set. This fact could be connected with the interference between communication and feedback on errors mechanisms, whose importance is discussed also in psychology [5-7].

Compared to the classical Backpropagation algorithm, the proposed approach is much more resource intensive, especially because both positive and negative examples has been taken into consideration. However, the amount of data stored into the ANN weights is augmented too, as it includes also language information. The "language-driven" Backpropagation algorithm could have also a cultural role, as words injected from the environment have an effect of modifying the distance between input vectors, by dividing the extended input space in sub-spaces sharing the same language component.

Further analysis along applying this approach to more complex scenarios is currently in progress, as we consider the proposed "language-driven" algorithm might contribute to a better understanding of learning in humans, opening as well the possibility of creating a specific category of artificial neural networks, with abstraction capabilities.